\documentclass[letterpaper,10pt,conference]{ieeeconf}

\IEEEoverridecommandlockouts
\usepackage{amsmath}
\usepackage{amssymb}
\usepackage{graphicx}
\usepackage{booktabs}
\usepackage{array}
\usepackage{cite}
\usepackage{flushend}
\usepackage[hidelinks]{hyperref}
\hypersetup{
    pdftitle={Single-View RGB-D Force--Deformation Estimation for Robotic Manipulation of Elasto-Plastic Materials},
    pdfauthor={Dezhi Sun, Marijn van Hilten, and Kenan Niu},
    pdfsubject={Corresponding author: Kenan Niu}
}

\title{\LARGE \bf
Single-View RGB-D Force--Deformation Estimation for Robotic Manipulation of Elasto-Plastic Materials
}

\author{Dezhi Sun, Marijn van Hilten, and Kenan Niu$^{*}$%
\thanks{The authors are with Robotics and Mechatronics, Faculty of Electrical
Engineering, Mathematics and Computer Science, University of Twente,
Enschede, The Netherlands.}%
\thanks{$^{*}$Corresponding author: Kenan Niu.\protect\newline \textcopyright~2026 IEEE. Personal use of this material is permitted. Permission from IEEE must be obtained for all other uses, in any current or future media, including reprinting/republishing this material for advertising or promotional purposes, creating new collective works, for resale or redistribution to servers or lists, or reuse of any copyrighted component of this work in other works.}%
}

\begin{document}
\maketitle
\thispagestyle{empty}
\pagestyle{empty}

\begin{abstract}
Persistent deformation complicates elasto-plastic manipulation by
coupling physical interaction with spatially distributed shape changes.
We investigate the force--deformation relationship using a standard
robot, a single eye-in-hand RGB-D camera, and intrinsic force estimates.
A 7-DoF Franka Emika FR3 performs normal indentations while an Intel
RealSense L515 captures pre- and post-contact surfaces. Point-cloud
processing localizes deformation and estimates displaced volume.
Force--volume characterization includes 27 measurements over nine
force levels. A compact multilayer perceptron (MLP), trained on
52 force--volume pairs at four levels, achieves a held-out RMSE of
1.857~N and $R^2=0.440$. A separate comparison shows lower error
for a quadratic model at 16~N and for the MLP at 19 and 23.5~N.
An eight-step experiment at 4~N per interaction demonstrates
cumulative deformation localization from the second indentation onward.
These results establish a practical measurement basis for
deformation-aware manipulation using intrinsic force estimates
and single-view geometry.
\end{abstract}

\begin{keywords}
Deformable object manipulation, elasto-plastic materials, force--deformation estimation, point clouds, RGB-D perception.
\end{keywords}

\section{Introduction}
Robotic manipulation of deformable objects is challenging because each interaction changes both object geometry and subsequent dynamics. For elasto-plastic materials, the deformation persists and is spatially distributed, while the observable response depends on surface state, interaction history, and material condition. A robot must therefore connect a commanded physical interaction to the resulting surface change without assuming rigid-body behavior~\cite{arriola-rios_modeling_2020,zhu_challenges_2022,gu_survey_2023}.

Recent systems have advanced geometric shaping of clay, dough, and related materials. RoboCraft learns particle dynamics for model-predictive shaping, and RoboCook extends learned interaction models to long-horizon tool use~\cite{shi_robocraft_2022,shi_robocook_2023}. SculptBot learns clay dynamics from point clouds, whereas SculptDiff learns goal-conditioned policies from demonstrations~\cite{bartsch_sculptbot_2023,bartsch_sculptdiff_2024}. PinchBot addresses long-horizon pinch actions with guided diffusion, and hierarchical text-to-3D shaping combines learned actions with high-level reasoning~\cite{bartsch_pinchbot_2025,bartsch_planning_2025}. Related approaches use compliant hardware, reinforcement learning, differentiable system identification, or learned predictive control~\cite{matl_deformable_2021,yang_differentiable_2025,zhang_manipulating_2025}. Dough shaping and robotic pizza stretching further illustrate task-specific sensing and action design~\cite{ondras_robotic_2022,kim_planning_2022}. These systems chiefly target goal geometry, learned dynamics, demonstrations, or policy learning.

Related studies have integrated robotic interaction with image-based
perception for ultrasound reconstruction \cite{sun2025arus},
force-controlled tissue segmentation \cite{sun2025hcce}, and
fluoroscopic guidance \cite{ward2025bilateral}.Sensing choices range from multimodal measurements to vision-only feedback~\cite{nadon_multi-modal_2018,zhang2026bci}. RGB-D probing has been used to track deformation and classify non-rigid response~\cite{hui_visual_2017}; visual shaping of plastic material has also been demonstrated with specialized actions and model-free image feedback~\cite{cherubini_towards_2018,cherubini_model-free_2020}. RoPotter instead exploits pottery-specific priors to reduce the state dimension and recover occluded geometry~\cite{yoo_ropotter_2024}. Fewer studies isolate a directly measured link between intrinsic interaction force and persistent geometric deformation from a single commodity view. We address this measurement problem with standard robotic hardware. The scope is controlled indentation of one Play-Doh-type material; closed-loop target-shape reproduction is outside the present evaluation.

We pair pre- and post-interaction observations, compute the estimated displaced volume from registered point clouds, and map it to force using a quadratic model and a compact multilayer perceptron (MLP). A separate sequence tests cumulative localization. The contributions are:
\begin{enumerate}
    \item a minimal-sensing real-robot interaction framework combining a 7-DoF robot, a single eye-in-hand RGB-D camera, and intrinsic force estimation for controlled interaction with an elasto-plastic material;
    \item a point-cloud deformation pipeline that localizes the affected surface region and computes an estimated deformation volume relative to a local reference surface; and
    \item a real-robot evaluation covering force–volume characterization, analytical and MLP-based force estimation, system repeatability, and cumulative deformation localization across sequential interactions.
\end{enumerate}

\begin{figure*}[t]
    \centering
    \begin{minipage}[c]{0.405\textwidth}
        \centering
        \includegraphics[width=\linewidth]{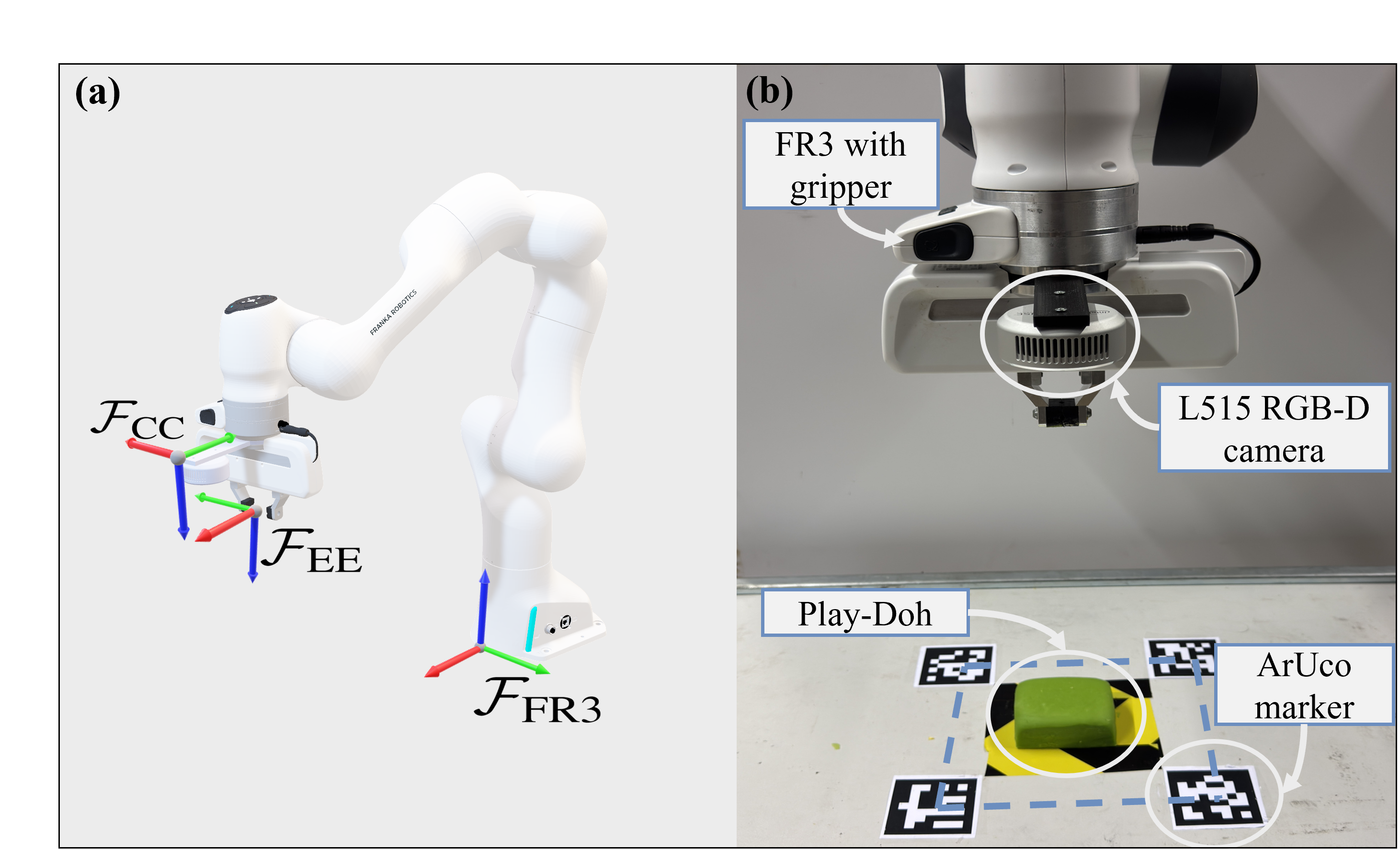}
    \end{minipage}\hfill
    \begin{minipage}[c]{0.575\textwidth}
        \centering
        \includegraphics[width=\linewidth]{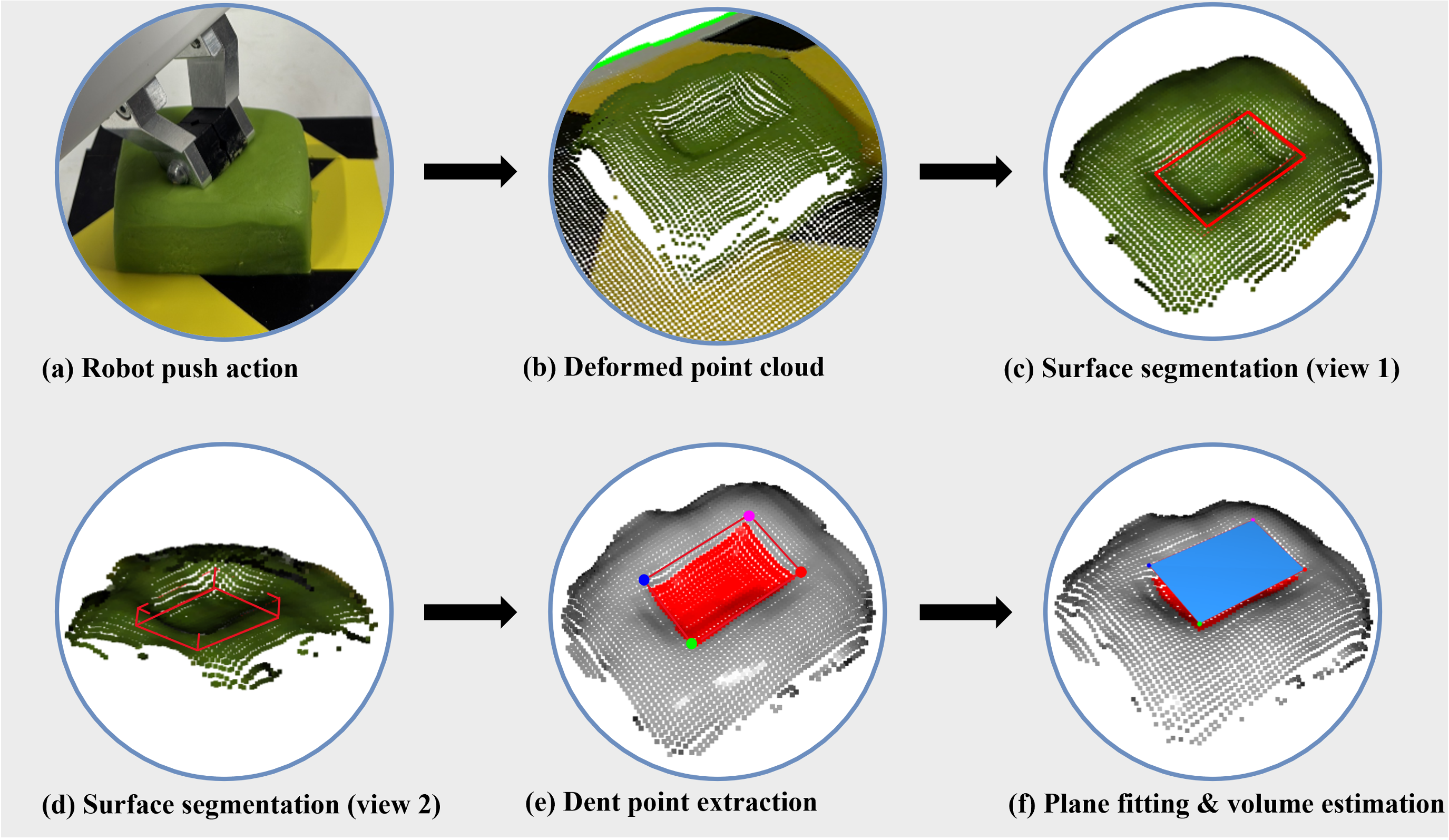}
    \end{minipage}
    \caption{System and deformation-estimation workflow.
Left: Robotic platform comprising the FR3, an eye-in-hand L515,
a Play-Doh-type specimen, and an ArUco-defined region of interest.
Right: Deformation-estimation pipeline comprising controlled
indentation, RGB-D acquisition, surface segmentation, deformation
extraction, reference-plane fitting, and estimated displaced-volume
computation.}
    \label{fig:system_pipeline}
\end{figure*}

\section{System and Method}
Figure~\ref{fig:system_pipeline} summarizes the measurement sequence: acquire the pre-interaction surface, indent the material, acquire the post-interaction surface, extract the affected region, compute estimated displaced volume, and relate it to interaction force.

The architecture couples two complementary measurements at different phases of an interaction. Intrinsic force is available during contact and terminates the indentation, whereas the eye-in-hand camera observes persistent geometry after the tool withdraws. This arrangement uses the manipulator and camera already present in many robotic workcells while retaining a direct connection between physical interaction and the resulting surface change.

\subsection{Minimal-Sensing Robotic Platform}
The platform comprises a 7-DoF Franka Emika FR3 and a LiDAR-based Intel RealSense L515 RGB-D camera rigidly attached to the end-effector. Four ArUco markers define the region of interest (ROI) around the specimen. Custom C++ and Python components use the Franka Control Interface and ROS~1 drivers~\cite{franka_robotics_gmbh_libfranka_2025,franka_robotics_gmbh_franka_ros_2025,intel_corporation_realsense-ros_2025} on Ubuntu~20.04 to coordinate motion, scanning, and indentation. The FR3 estimates end-effector force from measured joint torques and its internal robot model, avoiding an external force sensor.

The transformation chain contains the robot-base, end-effector, camera-color, and camera-depth frames. Eye-in-hand calibration uses a ChArUco board and the Tsai--Lenz method~\cite{tsai_real_1988,garrido-jurado_automatic_2014}; related formulations solve the robot--sensor transformation on the Euclidean group, and ROS tooling automates pose collection and solver execution~\cite{park_robot_1994,esposito_easy_handeye_2025}. With homogeneous transforms, a depth-camera point $\mathbf{p}_{\mathrm{CD}}$ is mapped to the robot base as
\begin{equation}
\mathbf{p}_{\mathrm{B}}=
\mathbf{H}_{\mathrm{B}\leftarrow\mathrm{EE}}
\mathbf{H}_{\mathrm{EE}\leftarrow\mathrm{CC}}
\mathbf{H}_{\mathrm{CC}\leftarrow\mathrm{CD}}
\mathbf{p}_{\mathrm{CD}}.
\label{eq:transform}
\end{equation}
Calibration repeatability is evaluated from 25 observations of the center of a square marker arrangement: five lateral camera positions are repeated at five capture heights. At each height and for each axis, absolute deviations are computed relative to the mean center over the five lateral observations at that height. The deviation magnitude generally increases with camera--surface distance; the $x$ estimate varies most and the $y$ estimate is most stable. Because no independently surveyed ground truth is available, these measurements quantify localization repeatability across viewpoints rather than absolute calibration accuracy. The sensitivity to pose range and measurement noise is consistent with established hand--eye calibration behavior~\cite{enebuse_accuracy_2022}.
For height $h$, lateral pose $l$, and coordinate $a$, repeatability is the absolute deviation $|q_{h,l,a}-\bar q_{h,a}|$ from the five-pose mean at that height. This definition deliberately avoids an unsupported ground-truth error claim.

\subsection{Force-Controlled Interaction}
Contact interaction has been addressed through learning-based
compliant control \cite{wu2022compliant} and admittance control
for robotic scanning \cite{davoodi2023control}.Each trial starts at an observation pose, records the undeformed surface, executes a normal indentation, returns to the observation pose, and records the deformed surface. During contact, a moving average filters the intrinsic normal-force estimate. The 1~kHz Franka Control Interface cycle monitors the filtered force and end-effector position. Interaction stops when
\begin{equation}
\left|F^{*}-\bar{F}_{z}\right|\leq F_{\mathrm{tol}},
\label{eq:force_stop}
\end{equation}
where $F^{*}$ is the target normal force, $\bar{F}_{z}$ is the filtered estimate, and $F_{\mathrm{tol}}$ is the tolerance. This position--impedance strategy regulates the stopping force but does not guarantee an identical tool trajectory across trials.

At 100~Hz, an inner update adjusts translational stiffness and advances the vertical target in small increments if the end-effector motion stagnates before satisfying Eq.~\eqref{eq:force_stop}. The baseline translational stiffness is 1000~N/m and is adapted to maintain compliant contact while approaching the target. The controller therefore combines a force-based termination condition with position--impedance motion; the observed trajectory remains an experimental outcome rather than a prescribed force trajectory.

Returning to a common observation pose separates contact execution from geometric measurement. Each deformation estimate is computed after tool withdrawal from a paired pre-contact and post-contact observation, so the measured surface change represents persistent deformation rather than instantaneous tool occlusion. The recorded trajectories retain the physical path taken to the common stopping-force condition and are used separately to assess interaction repeatability.

\subsection{Vision-Based Deformation Estimation}
The vision pipeline uses Open3D~\cite{zhou_open3d_2018} and has distinct single-indentation and sequential processing modes.

\textit{Single-indentation volume estimation:} A pre-interaction source cloud and post-interaction target cloud are cropped to the marker-defined ROI. RANSAC plane fitting removes background points~\cite{fischler_random_1981}, and neighbor-based filtering removes isolated measurements. The known interaction location and tool footprint define a local support region in which a reference plane is fitted. For the $N_b$ target points below that plane, the represented area per point is $A_p=A_c/N_b$, where $A_c$ is the contact/reference area in mm$^2$. If $z_i$ and $z_{\mathrm{plane},i}$ are the measured and reference-plane heights in mm, respectively, estimated displaced volume is
\begin{equation}
V=\sum_{i=1}^{N_b}A_p\left(z_{\mathrm{plane},i}-z_i\right),
\label{eq:volume}
\end{equation}
with $V$ in mm$^3$.

The scalar $V$ is a geometric estimate of the visible surface depression relative to a locally fitted reference plane. It is used as a repeatable interaction descriptor rather than as a direct measurement of conserved material volume.

\textit{Sequential deformation localization:} Consecutive source and target clouds undergo the same ROI segmentation and outlier removal. Surface normals are estimated and consistently oriented, after which point-to-plane iterative closest point (ICP) registration aligns the target to the source~\cite{rusinkiewicz_efficient_2001}. For each target point $\mathbf{p}$, the $k$ nearest source points define a local centroid $\mathbf{c}$ and averaged unit normal $\hat{\mathbf{n}}$. Their signed surface distance is
\begin{equation}
d=(\mathbf{p}-\mathbf{c})\cdot\hat{\mathbf{n}}.
\label{eq:signed_distance}
\end{equation}
A median-absolute-deviation threshold identifies the lower-tail displaced points~\cite{hampel_influence_1974}. The sign of $d$ follows the consistently oriented source normals; its magnitude expresses local surface displacement, while the retained sign distinguishes motion along or opposite the normal. DBSCAN groups thresholded points using $\varepsilon=2\Delta$, where $\Delta$ is their mean point spacing~\cite{ester_density-based_1996}. For cluster $C_i$, the deformation score is $s_i=-P_{20}(d(C_i))$, where $P_{20}$ is the 20th percentile of its signed distances; $\arg\max_i s_i$ is retained as the dominant affected region. An oriented bounding box defines local support for reference-plane fitting, and Eq.~\eqref{eq:volume} estimates cumulative volume. This mode does not assume a known individual press location and does not identify or count neighboring dents separately.

The two modes serve complementary measurement tasks. Single-indentation processing uses the known interaction site and tool footprint to obtain a local volume estimate for force regression. Sequential processing instead aligns consecutive observations and selects the dominant displaced cluster without requiring the location of each press as input. Both modes reduce the observed surface change to geometric quantities that can be associated with the robot's intrinsic force estimate.

\subsection{Force--Volume Regression}
The analytical baseline is a quadratic fitted to 27 single-indentation measurements. With force $F$ in N and estimated displaced volume $V$ in mm$^3$, it is
\begin{equation}
F=-9.377\times10^{-8}V^2+0.00266V+5.567.
\label{eq:polynomial}
\end{equation}

The compact nonlinear comparator is a fully connected 1--32--32--1 MLP implemented in PyTorch. Each hidden layer uses layer normalization and ReLU, followed by dropout of 0.1. Training uses Huber loss, AdamW with learning rate $10^{-3}$, ReduceLROnPlateau, batch size 8, a maximum of 1000 epochs, and early stopping with patience 30. Inputs and outputs are min--max scaled and converted back to physical units for evaluation. Both regressors use the volume estimate as their only predictor.

For $n$ held-out observations, performance is reported in physical force units using
\begin{equation}
\begin{aligned}
\mathrm{RMSE}&=\sqrt{\frac{1}{n}\sum_{i=1}^{n}(F_i-\hat F_i)^2},\\
R^2&=1-\frac{\sum_i(F_i-\hat F_i)^2}{\sum_i(F_i-\bar F)^2}.
\end{aligned}
\label{eq:metrics}
\end{equation}
The separate three-level comparison computes per-force RMSE from identical volume inputs for both models. Both models are evaluated as geometry-to-force estimators, quantifying how much information about interaction force remains in the post-contact surface deformation. Their use for closed-loop force selection is outside the present evaluation.

\section{Experimental Protocol}
Table~\ref{tab:datasets} summarizes six complementary protocols used for force--volume characterization, model evaluation, robot-motion repeatability, camera localization repeatability, and sequential cumulative-deformation analysis.

The characterization, MLP, and common-input comparison sets serve distinct modeling roles. The trajectory, camera, and sequential protocols evaluate complementary parts of the physical measurement process.

Each protocol answers a separate system question. Characterization establishes the measured force--volume trend, the grouped split tests geometry-to-force prediction, and the comparison set evaluates both regressors on identical inputs at three additional forces. The trajectory and camera protocols quantify repeatability in contact execution and geometric localization, respectively. The sequential experiment then tests whether the vision pipeline can follow an affected region as adjacent interactions accumulate.

\subsection{Force--Volume Characterization}
The robot performs three single indentations at each of nine force levels from 5 to 25~N in 2.5~N increments, producing 27 force--volume measurements. Equation~\eqref{eq:polynomial} is fitted to this characterization set.

\subsection{Force-Prediction Evaluation}
The MLP dataset contains 52 pairs at 15, 17.5, 20, and 22.5~N, with 13 pairs per level. Samples are grouped by session and force level followed by a 70\%/15\%/15\% training, validation, and test split. Held-out predictions are evaluated using RMSE and $R^2$. A separate comparison uses three volume observations at each of 16, 19, and 23.5~N as common inputs to both regressors. One hundred training runs assess neural-network initialization sensitivity under the fixed split.

\subsection{Calibration and Interaction Repeatability}
The camera-to-robot localization study uses 25 observations arranged as five lateral camera positions at each of five capture heights. Interaction repeatability is tested in five trials with nominally identical Play-Doh pieces, setup, controller parameters, and a 16~N stopping-force target. End-effector $x$, $y$, and $z$ trajectories are recorded at 1~kHz; downsampling to 5~Hz is used only for visualization and does not change the controller or endpoint statistics.

\subsection{Sequential Deformation Experiment}
An eight-step sequential experiment applies eight adjacent circular indentations along a straight line with a custom tool at 4~N per interaction. One point cloud is acquired after each indentation. The experiment evaluates cumulative affected-region localization rather than individual dent identification.

\begin{table}[t]
\centering
\caption{Experimental protocols and evaluation roles.}
\label{tab:datasets}
\setlength{\tabcolsep}{2pt}
\renewcommand{\arraystretch}{1.05}
\begin{tabular}{@{}>{\raggedright\arraybackslash}p{0.27\columnwidth}>{\raggedright\arraybackslash}p{0.27\columnwidth}>{\raggedright\arraybackslash}p{0.37\columnwidth}@{}}
\toprule
Protocol & Design & Evaluation role \\
\midrule
Characterization & 5--25~N, 2.5~N steps; 3/level (27) & Force--volume response and quadratic fit \\
MLP & 15, 17.5, 20, 22.5~N; 13/level (52) & Grouped 70/15/15 train/validation/test split \\
Comparison & 16, 19, 23.5~N; 3/level (9) & Common inputs for both regressors \\
Trajectory & 16~N; five trials & End-effector repeatability \\
Sequential & One eight-step sequence; 4~N/step & Cumulative deformation localization \\
Camera & 5 poses at 5 heights (25) & Localization repeatability \\
\bottomrule
\end{tabular}
\end{table}

\begin{figure}[t]
    \centering
    \includegraphics[width=\columnwidth]{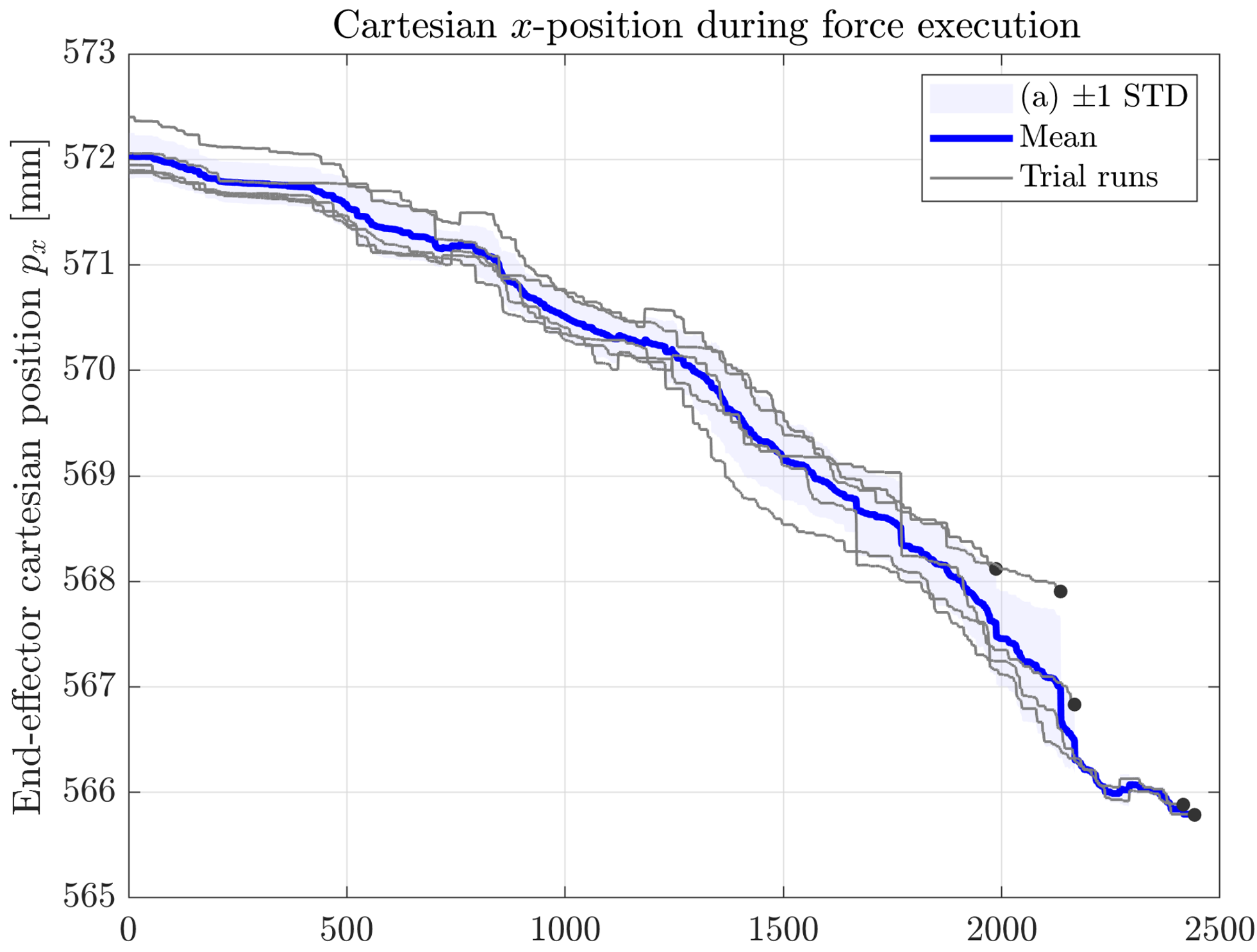}
    \caption{End-effector $x$ position during five nominally identical 16~N trials. The trajectories show systematic lateral drift despite the common stopping-force target; downsampling from 1~kHz to 5~Hz is for visualization only.}
    \label{fig:force_repeatability}
\end{figure}

\begin{figure*}[t]
    \centering
    \begin{minipage}[t]{0.315\textwidth}
        \centering
        \includegraphics[width=\linewidth,trim=0 0 0 90,clip]{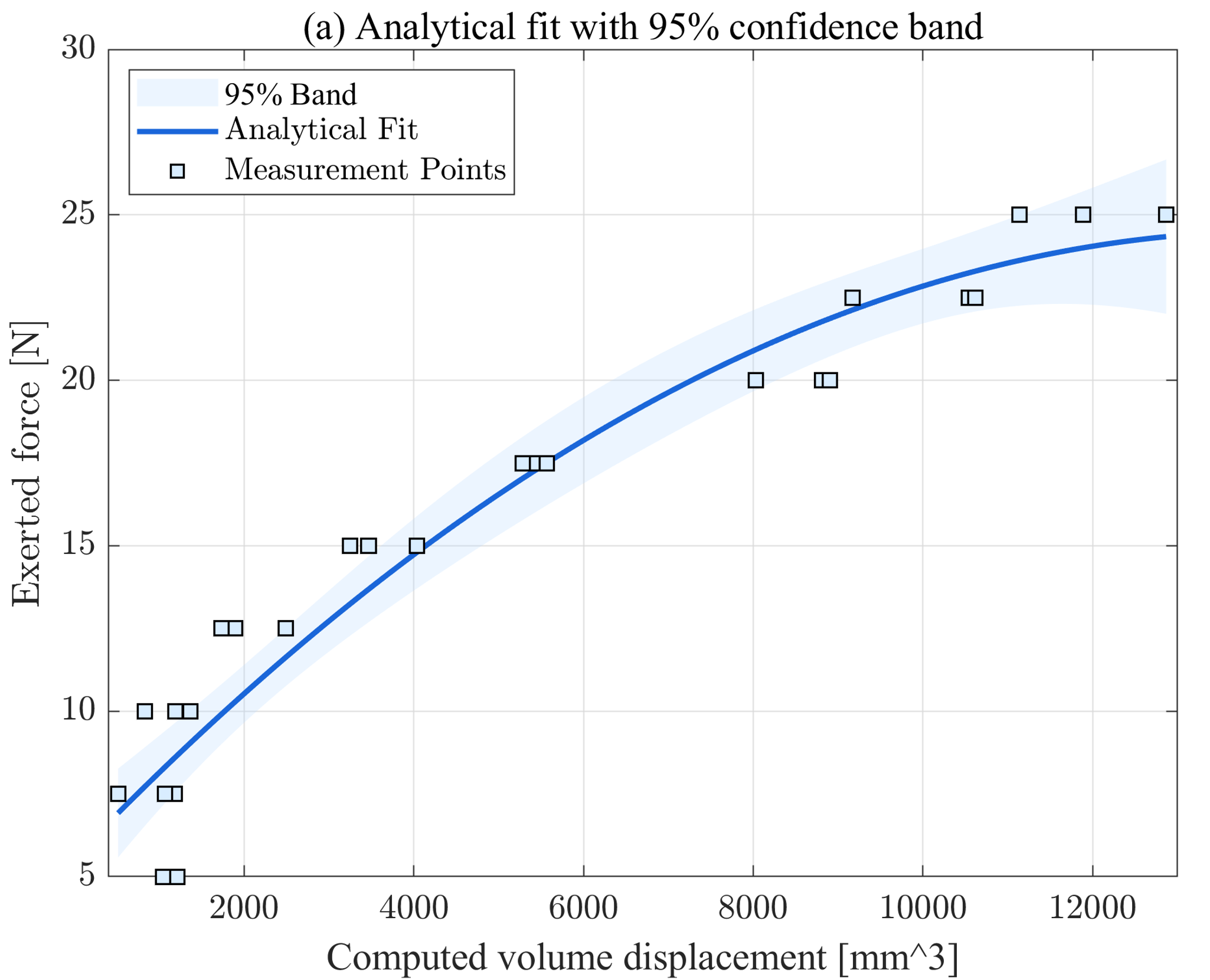}\\[-1mm]
        \footnotesize (a)
    \end{minipage}\hfill
    \begin{minipage}[t]{0.315\textwidth}
        \centering
        \includegraphics[width=\linewidth,trim=0 0 0 90,clip]{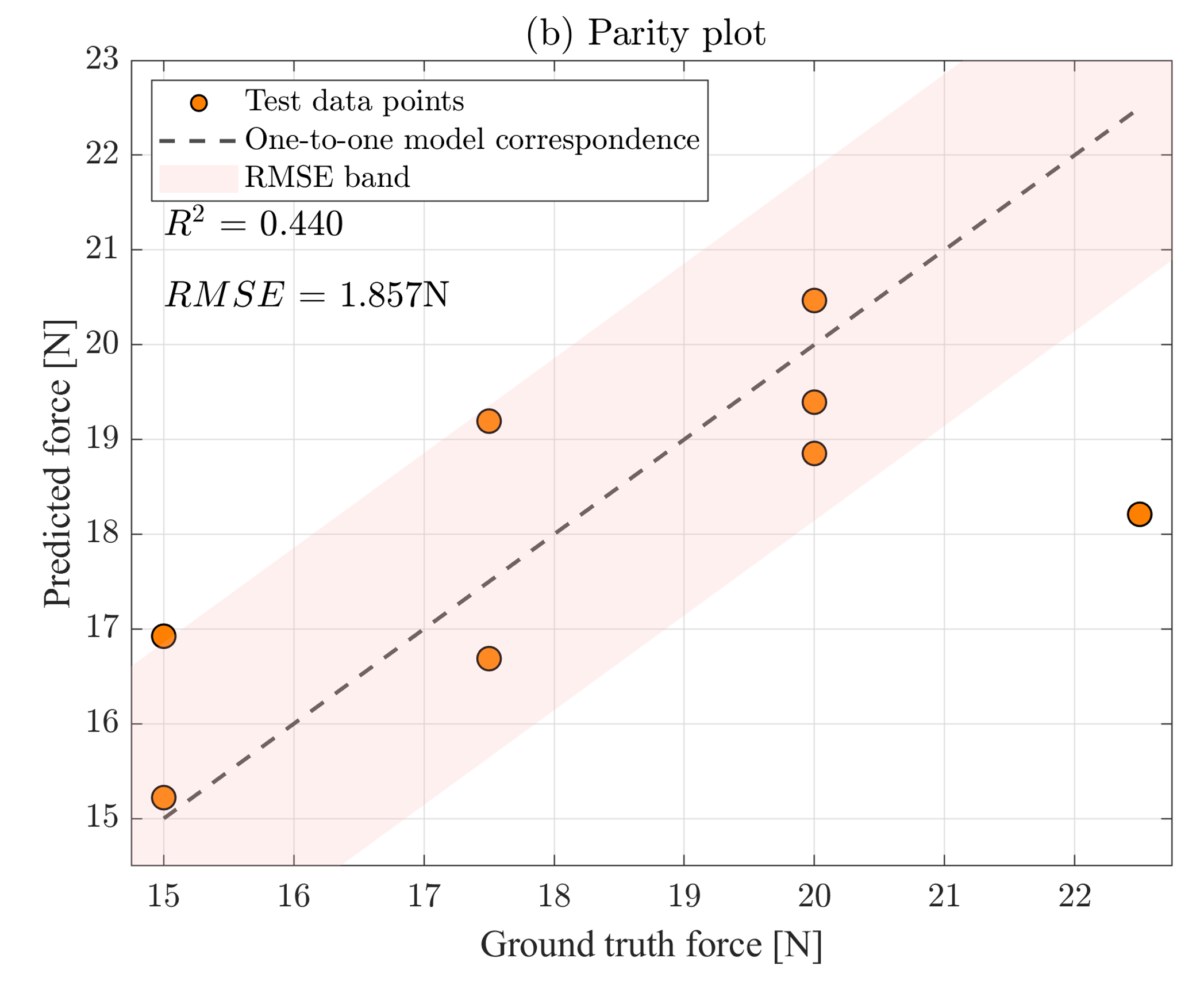}\\[-1mm]
        \footnotesize (b)
    \end{minipage}\hfill
    \begin{minipage}[t]{0.315\textwidth}
        \centering
        \includegraphics[width=\linewidth,trim=0 0 0 90,clip]{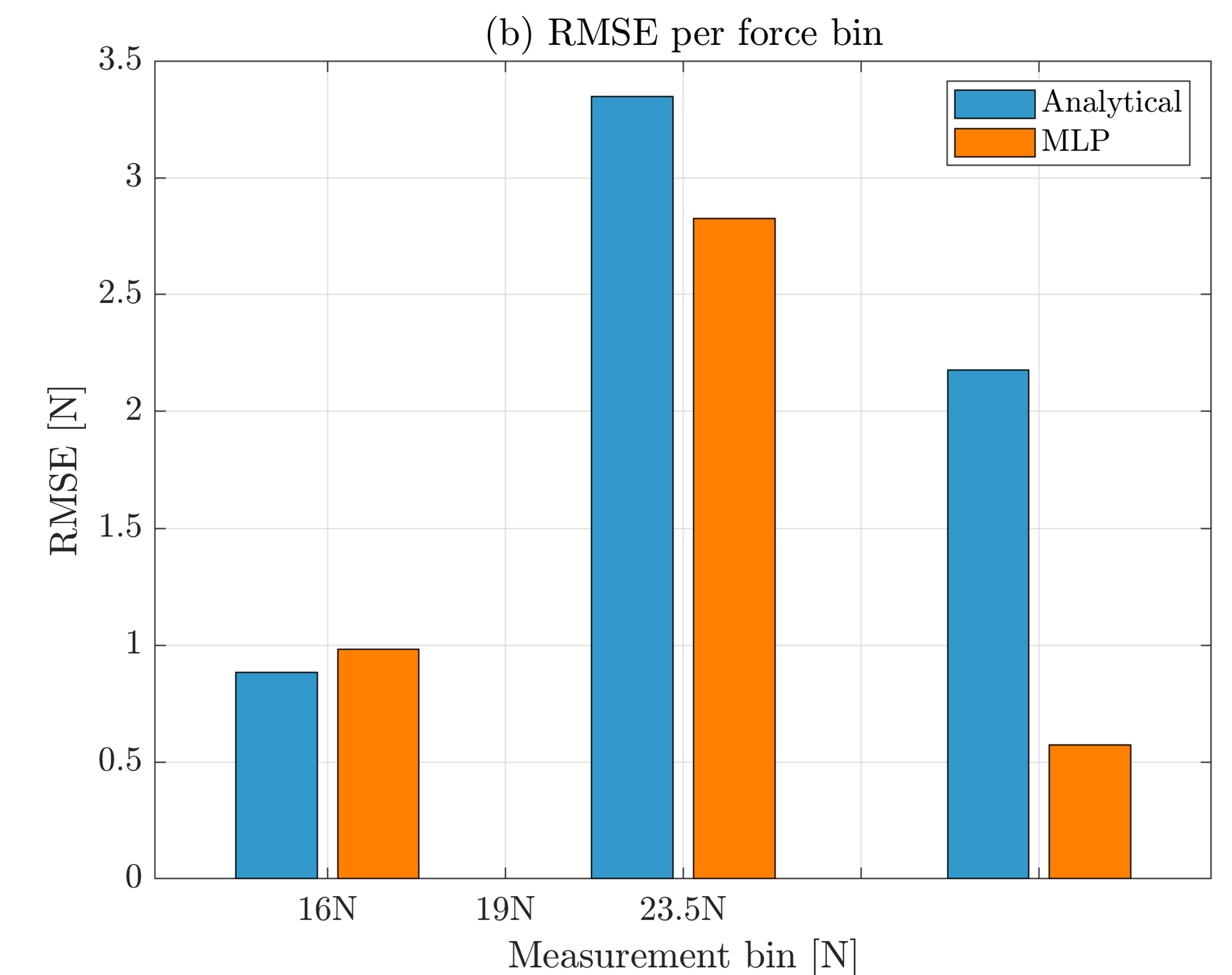}\\[-1mm]
        \footnotesize (c)
    \end{minipage}
    \caption{Force--volume and prediction results. (a) Measured force versus estimated displaced volume, quadratic fit, and reported 95\% prediction band. (b) MLP predictions on the held-out test subset. (c) Per-force RMSE on the separate comparison set; each force level contains three observations evaluated by both models.}
    \label{fig:quantitative}
\end{figure*}

\section{Results and Discussion}
\subsection{Force--Volume Characterization}
Figure~\ref{fig:quantitative}(a) shows limited distinguishable volume displacement at forces up to approximately 10~N relative to background surface variation. Measurable displacement becomes clearer from approximately 12.5~N and generally increases with force. Variability widens at higher levels, and the measured relation becomes more nonlinear toward the upper end of the tested range. These observations characterize the measured response and are not interpreted as verified constitutive material phases.

The characterization therefore identifies an experimentally useful force range where persistent surface displacement is distinguishable from the measured background variation. The quadratic fit summarizes the full 5--25~N response with a compact analytical mapping, while the prediction band visualizes the observed spread around that mapping. This characterization supplies both a baseline estimator and the force–volume context for interpreting the MLP results.

\subsection{Analytical and Learning-Based Prediction}
On the held-out subset, the MLP obtains an RMSE of 1.857~N and $R^2=0.440$ [Fig.~\ref{fig:quantitative}(b)]. The MLP captures the overall nonlinear force--volume trend, while the residual errors indicate that estimated displaced volume alone does not encode all interaction variability. The 22.5~N held-out sample shows notable under-prediction.

The separate model comparison gives a conditional rather than universal ranking [Fig.~\ref{fig:quantitative}(c)]. The analytical model has slightly lower error at 16~N, whereas the MLP has lower error at 19 and 23.5~N; both models show their largest reported error at 19~N. The 100 runs evaluate initialization sensitivity under the fixed data split and show consistent convergence behavior.

Together, the two estimators provide complementary representations of the measured relationship. The quadratic model offers a closed-form description across the characterization range, while the MLP captures nonlinear variation from the four-level training set. Evaluating both on identical comparison inputs makes their force-dependent behavior directly visible without reducing the result to a single aggregate ranking.

\begin{figure*}[t]
    \centering
    \begin{minipage}[t]{0.315\textwidth}
        \centering
        \includegraphics[width=\linewidth,trim=0 155 0 0,clip]{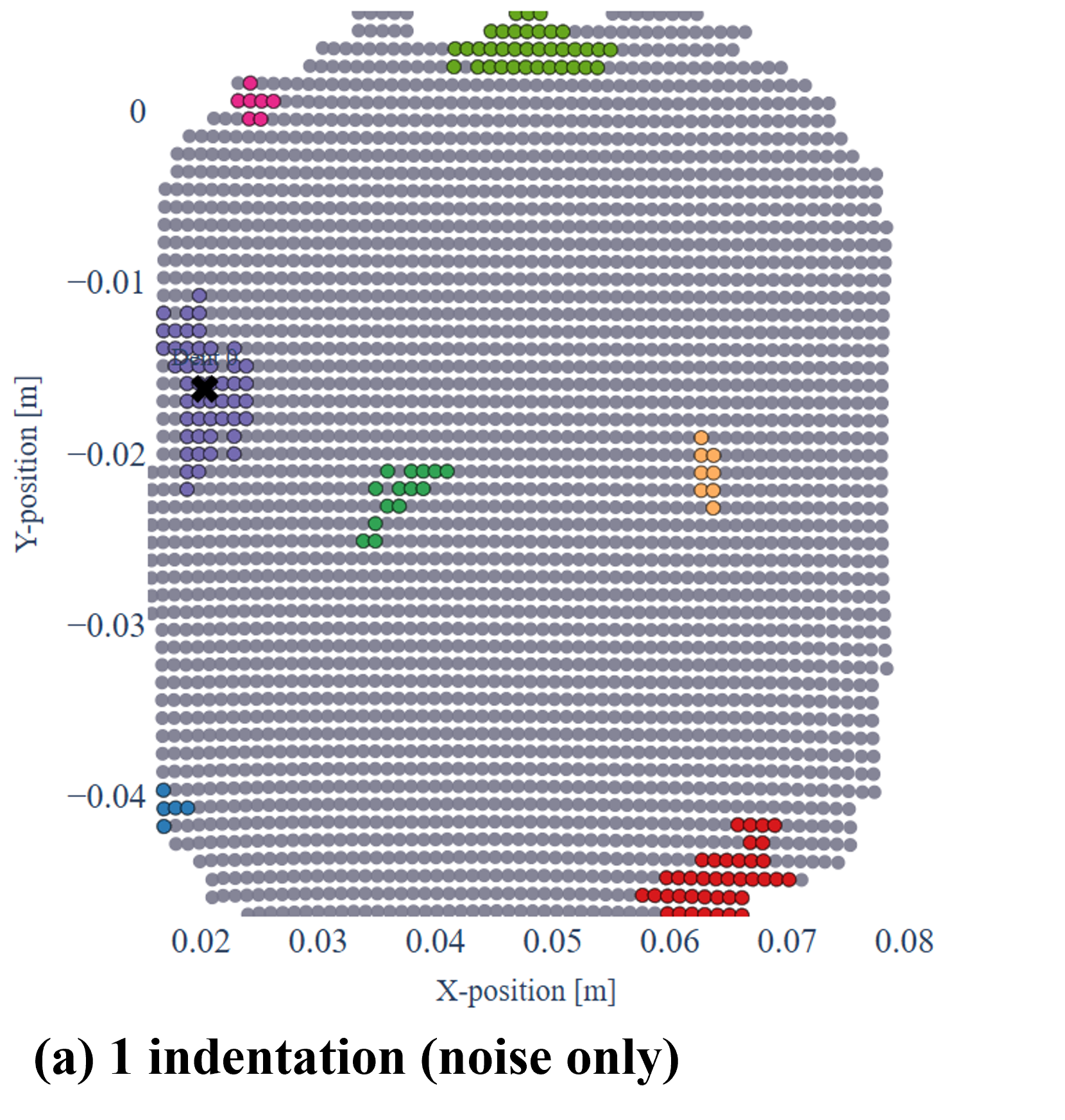}\\[-1mm]
        \footnotesize (a) One indentation
    \end{minipage}\hfill
    \begin{minipage}[t]{0.315\textwidth}
        \centering
        \includegraphics[width=\linewidth,trim=0 155 0 0,clip]{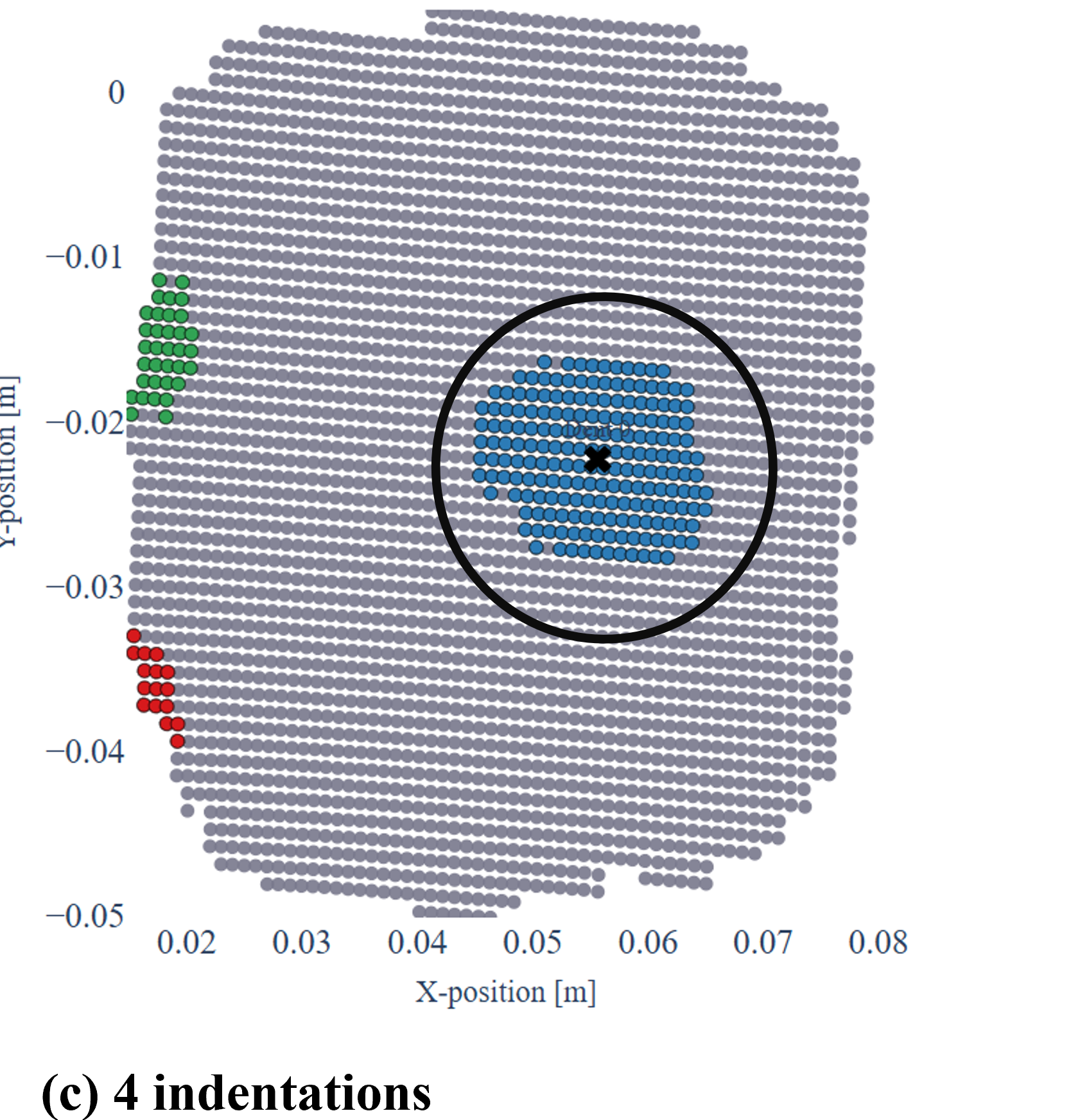}\\[-1mm]
        \footnotesize (b) Four indentations
    \end{minipage}\hfill
    \begin{minipage}[t]{0.315\textwidth}
        \centering
        \includegraphics[width=\linewidth,trim=0 155 0 0,clip]{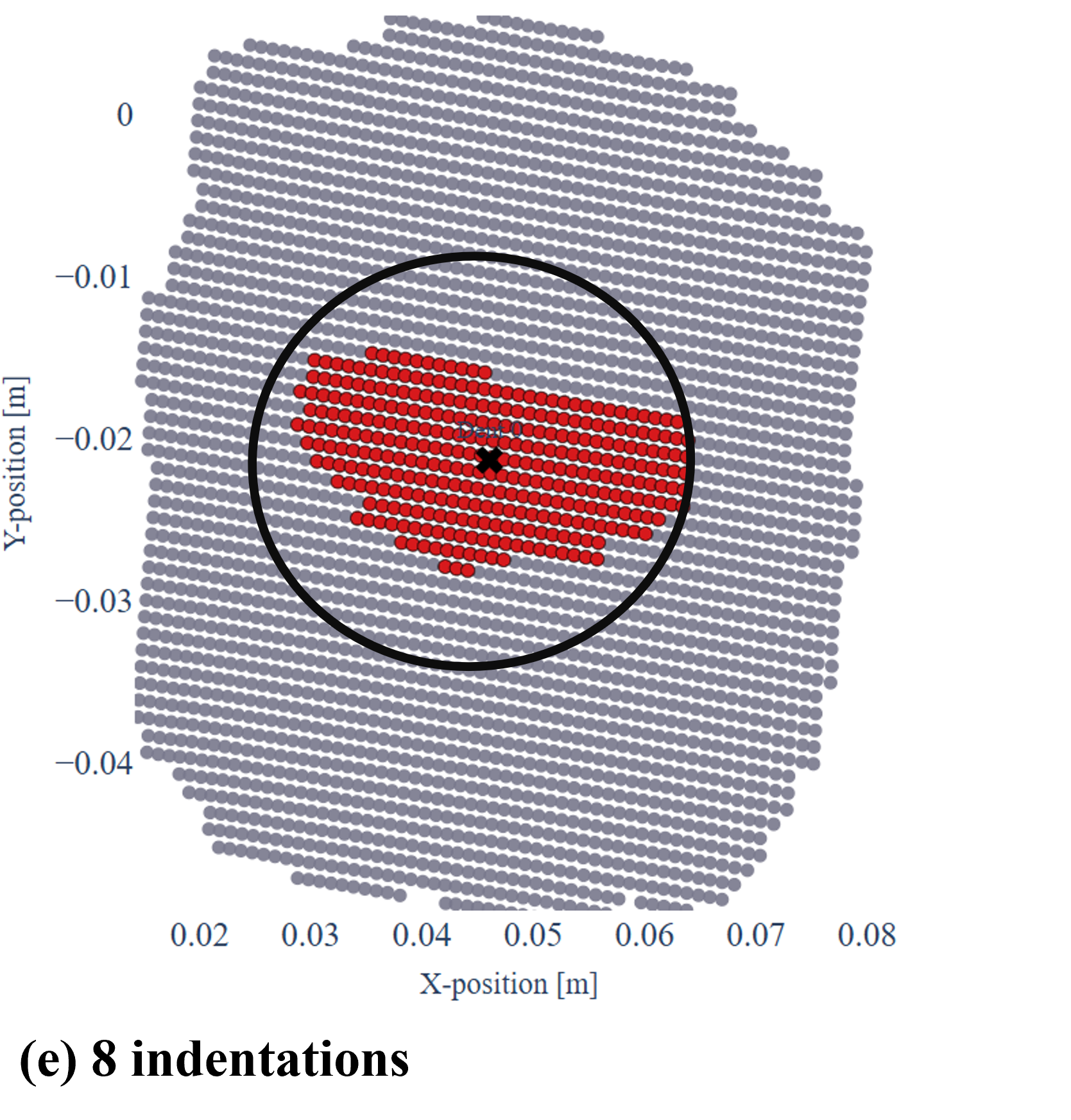}\\[-1mm]
        \footnotesize (c) Eight indentations
    \end{minipage}
    \caption{Sequential cumulative-deformation localization after one, four, and eight adjacent interactions. Gray points represent retained surface samples; colors indicate DBSCAN cluster identity rather than displacement magnitude. The black outline marks the selected dominant region, and the cross indicates its centroid.}
    \label{fig:sequential}
\end{figure*}

\subsection{System Repeatability and Error Sources}
The five 16~N trials show that a common stopping-force target can be reached through different contact paths. The $x$ trajectories vary by up to approximately 6~mm [Fig.~\ref{fig:force_repeatability}], whereas within-trial variation in $y$ is smaller. Final penetration depth in $z$ varies by approximately 5~mm across trials. These path differences affect the location and geometry of the resulting deformation and therefore contribute to force--volume variability.

The 25-pose localization study likewise shows that repeatability degrades as camera--surface distance increases. Variation is largest along $x$ and smallest along $y$. This is not an absolute-accuracy result, but it isolates viewpoint-dependent localization as another contributor to ROI placement, registration, and volume-estimation uncertainty.

The two repeatability tests expose different parts of the measurement chain. Camera localization changes the inferred surface coordinates before volume computation, whereas trajectory variation changes the physical contact itself. Their effects accumulate when a laterally shifted indentation is observed through a viewpoint-dependent transform and summarized by the volume estimate.

\subsection{Sequential Deformation Localization}
The first shallow deformation is not reliably separated from background surface variation [Fig.~\ref{fig:sequential}(a)]. From the second interaction onward, a dominant affected region emerges and expands along the interaction path. As neighboring dents merge, the algorithm tracks the cumulative affected region rather than preserving individual dent identity. The selected cluster provides a compact region, centroid, and cumulative volume estimate for subsequent interaction planning.

\subsection{Design Implications}
The results support a staged deformation-aware manipulation strategy in which intrinsic force terminates the current contact and post-contact geometry informs the next action. The RGB-D observation is acquired after the tool returns to the observation pose, avoiding tool occlusion while measuring persistent deformation. The present work evaluates the underlying measurement relation rather than a complete closed-loop controller.

The conditional model comparison argues against selecting a regressor from one aggregate score alone. The quadratic model is compact and interpretable inside the characterized range, whereas the MLP provides a compact nonlinear alternative. The force-wise results show how these complementary estimators can characterize the measured relation before integration with action selection.

\subsection{Limitations}
The evaluation focuses on one Play-Doh-type material under controlled laboratory conditions. Material condition, RGB-D surface noise, hand--eye localization variation, and intrinsic force estimation contribute to measurement variability. The single-view volume approximation does not recover self-occluded or laterally displaced material, and adjacent deformation regions may merge during sequential interaction. The MLP is evaluated over a limited force range, while broader material transfer, session-level generalization, and closed-loop target-shape reproduction remain topics for future evaluation.

\section{Conclusion}
This paper linked intrinsic interaction force to persistent deformation measured by a single eye-in-hand RGB-D camera. Indentation experiments revealed a nonlinear force--volume relationship, with a compact MLP achieving a held-out RMSE of 1.857~N and $R^2 = 0.440$. Sequential trials demonstrated cumulative deformation localization, supporting deformation-aware manipulation with standard robotic hardware.

\sloppy
\bibliographystyle{IEEEtran}
\bibliography{robio_references}

\end{document}